\documentclass[letterpaper, 10 pt, conference]{ieeeconf}  % Comment this line out if you need a4paper
\usepackage{graphicx}
\usepackage{cite}
\usepackage{color}
\usepackage{amsmath}
\usepackage{amsfonts}

\IEEEoverridecommandlockouts                              % This command is only needed if 
\title{\LARGE \bf
Verti-Arena: A Controllable and Standardized Indoor Testbed for Multi-Terrain Off-Road Autonomy
}

\author{
Haiyue Chen\textsuperscript{1}, 
Aniket Datar\textsuperscript{2}, 
Tong Xu\textsuperscript{2}, 
Francesco Cancelliere\textsuperscript{3},
Harsh Rangwala\textsuperscript{2}, \\
Madhan Balaji Rao\textsuperscript{2}, 
Daeun Song\textsuperscript{2},
David Eichinger\textsuperscript{2}, 
and Xuesu Xiao\textsuperscript{2}%
\thanks{\textsuperscript{1}University of Pennsylvania, Philadelphia, PA 19104, USA. \texttt{haiyuech@seas.upenn.edu}}%
\thanks{\textsuperscript{2}George Mason University, Fairfax, VA 22030, USA. \texttt{\{adatar, txu25, hrangwa, mbalajir, dsong26, deiching, xiao\}@gmu.edu}}%
\thanks{\textsuperscript{3}University of Catania, Catania, 95123, Italy. \texttt{francesco.cancelliere@phd.unict.it}}%
}

\begin{document}

\maketitle
\IEEEpeerreviewmaketitle

\thispagestyle{empty}
\pagestyle{empty}

%%%%%%%%%%%%%%%%%%%%%%%%%%%%%%%%%%%%%%%%%%%%%%%%%%%%%%%%%%%%%%%%%%%%%%%%%%%%%%%%
\begin{abstract}
Off-road navigation is an important capability for mobile robots deployed in environments that are inaccessible or dangerous to humans, such as disaster response or planetary exploration. Progress is limited due to the lack of a controllable and standardized real-world testbed for systematic data collection and validation. To fill this gap, we introduce Verti-Arena, a reconfigurable indoor facility designed specifically for off-road autonomy. By providing a repeatable benchmark environment, Verti-Arena supports reproducible experiments across a variety of vertically challenging terrains and provides precise ground truth measurements through onboard sensors and a motion capture system. Verti-Arena also supports consistent data collection and comparative evaluation of algorithms in off-road autonomy research. We also develop a web-based interface that enables research groups worldwide to remotely conduct standardized off-road autonomy experiments on Verti-Arena.

\end{abstract}

%%%%%%%%%%%%%%%%%%%%%%%%%%%%%%%%%%%%%%%%%%%%%%%%%%%%%%%%%%%%%%%%%%%%%%%%%%%%%%%%
\section{INTRODUCTION}
Autonomous off-road navigation enables rescue robots to enter ruins, jungles, and other disaster environments that are difficult or impossible for humans to access, and to perform search and rescue tasks. Researchers have shown that certain types of robots can traverse rubble, narrow crevices, dense vegetation, and coastal terrain to a limited extent~\cite{menna2014real, whitman2018snake, weerakoon2023vapor,xiao2017uav, tomic2012toward, michael2014collaborative}. One key focus in this area of research is developing wheeled robots that can traverse uneven surfaces and steep slopes while completing tasks across continuously varying terrains~\cite{datar2024toward, cai2025pietra, nazeri2024verticoder, xu2025verti}.

\par
Due to the complexity of off-road terrain, factors such as slope and tilt, surface roughness, and changes in friction can all affect the ability to achieve reliable mobility in off-road environments. In such cases, relying solely on the vehicle’s kinematic model is insufficient. It is necessary to consider complex kinodynamics that incorporates both environmental information and interactions between the vehicle and the terrain~\cite{ datar2024terrain, ward2008dynamic}. However, since the kinodynamics is largely affected by the environment, it becomes essential to perceive the environment, update the kinodynamic model accordingly, and plan based on this updated model~\cite{ward2025online,xiao2024foundation}.
\par
This approach requires data collected under realistic, physically diverse terrain conditions, where robot-environment interactions can be captured and used to model kinodynamics and evaluate off-road autonomy. 
However, the dynamics of diverse terrains are highly complex. Simulators often simplify the environment and approximate physical interactions, replacing real friction, wheel sinkage, and gravel rolling with idealized models. This limits their ability to support high-fidelity data collection, training, and testing. Collecting data in outdoor environments, on the other hand, faces challenges such as high operational cost, extended experiment time, and the lack of high-precision ground-truth data.
\par
\begin{figure}[t]
  \centering
\includegraphics[width=1.0\linewidth]{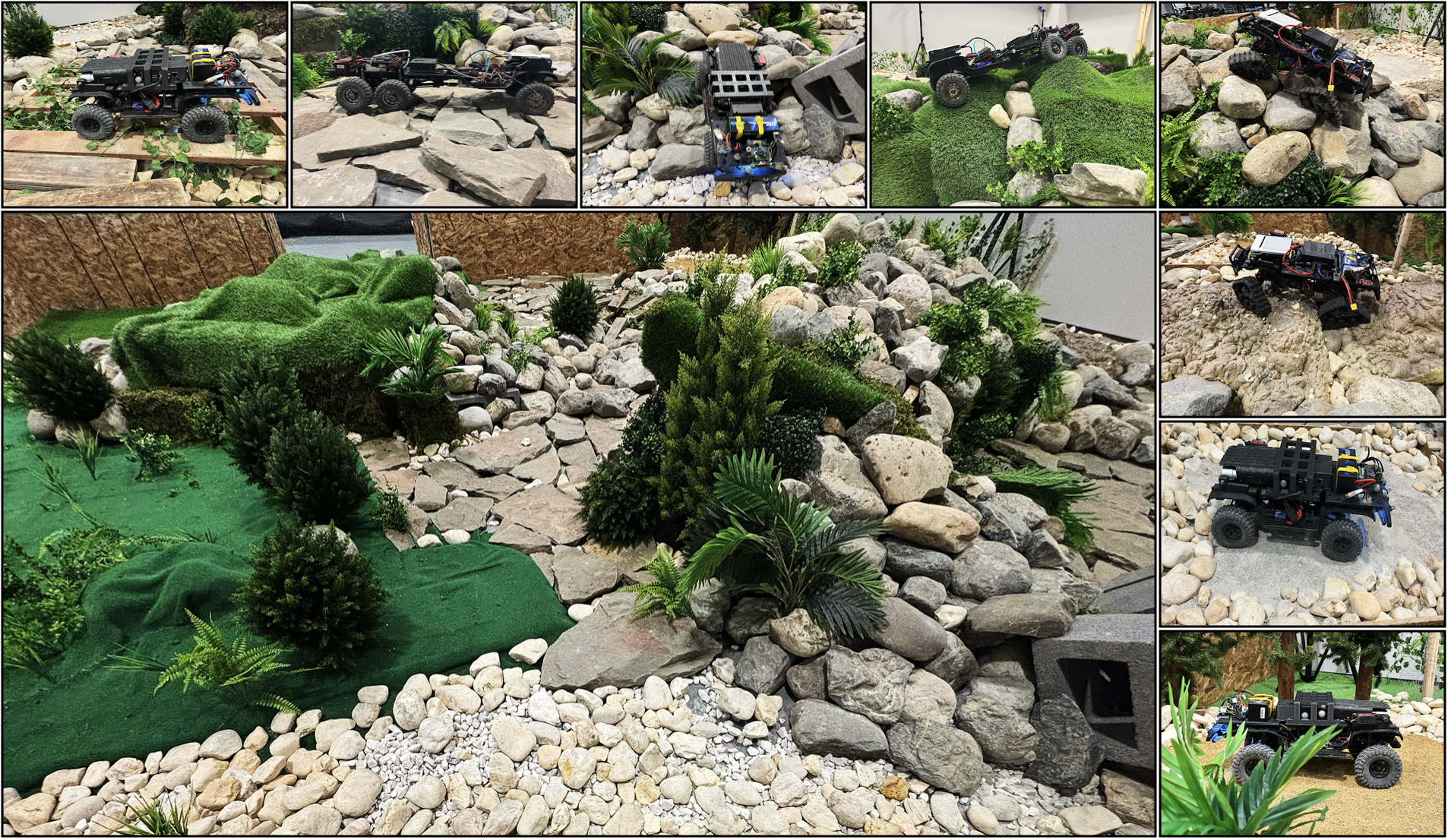}
  \caption{Verti-Arena comprises a variety of off-road terrain, includes different geometries and semantics, and is equipped with a motion capture system, to facilitate off-road autonomy research.}
  \label{fig:verti_arena}
\end{figure}

To address these limitations, Verti-Arena is a controllable and standardized indoor testbed featuring multi-terrain and vertically challenging conditions for off-road autonomy. Within a controlled laboratory setting, we combine ten types of terrain, including rocks, sand, grass, and other natural surfaces, within an 8$\times$8 m area featuring a maximum elevation difference of 0.7 meters (see Fig.~\ref{fig:verti_arena}). Precise ground-truth trajectories are provided using onboard sensors and a motion capture system. Verti-Arena enables the collection of large-scale datasets containing synchronized vehicle sensor data (e.g., RGB-D images and IMU measurements), control inputs, and accurate exteroception of robot poses, establishing a reliable and safe benchmark for future research in perception, control, and learning for off-road autonomy.

\section{RELATED WORK}
This section reviews physical test environments and existing datasets related to off-road autonomy.

\subsection{Physical Test Environments}
Several small indoor testbeds are designed to evaluate specific robotic capabilities. One series of testbeds, constructed from stacked rocks to create vertically challenging terrain, enables repeatable assessment of 1/10-scale four-wheeled or six-wheeled vehicles as they traverse steep inclines~\cite{datar2024toward, datar2024learning, datar2024terrain, gupta2025t}. Another indoor setup consists of a wooden floor with a gap filled with shredded paper, which is used to evaluate legged robots on collapsible footholds~\cite{tennakoon2020probe}. Robotarium~\cite{wilson2020robotarium}, a platform composed of 20 robots, is used to validate distributed control strategies in swarm robotics. SCATTER~\cite{qian2015dynamics} features boulders buried in sand and is designed to explore how spatial heterogeneity affects locomotion. The precisely controlled conditions of these indoor testbeds for specific robot skills ensure high repeatability.

Closed-course outdoor off-road tracks provide a more realistic but still controlled environment. However, because the terrain is relatively monotonous and the geometric features are simple, these tracks lack the complexity typically associated with true off-road environments. As a result, they are primarily used for high-speed driving tests~\cite{goldfain2019autorally,levy2025meta}.

Some studies turned to full-scale proving grounds~\cite{sivaprakasam2024tartandrive,talia2023demonstrating,triest2022tartandrive,triest2023learning,yu2024adaptive,li2025seb}, which include kilometer-long forest loops, ravines, and muddy terrain, to validate highly realistic vehicle dynamics. However, the high maintenance and operation cost prevents wide adoption of such expensive testbeds for many robotics researchers. Furthermore, the unpredictability introduced by outdoor weather and season conditions reduces the level of controllability and reproducibility in these environments. 
\par
Although existing testbeds have proven useful, they have struggled to balance environmental diversity with experimental controllability and repeatability. Verti-Arena introduces an indoor testbed that includes a broad range of seamlessly integrated vertically challenging terrain types while preserving the precise controllability found in laboratory settings.

\subsection{Datasets}
Many off-road autonomy datasets are collected in existing physical test environments. Several focus on perception challenges posed by conditions that are particularly difficult for reliable sensing. For instance, the M2P2 dataset~\cite{datar2024m2p2} emphasizes passive perception under extremely low light. DiTer++~\cite{kim2024diter++} uses multiple robotic platforms to collect multimodal terrain data. The GND dataset~\cite{liang2024gnd} adds passability classification to quantify traversal risk, enabling different robot types to assess navigability based on their capabilities. RELLIS 3D~\cite{jiang2021rellis}, the Great Outdoors Dataset~\cite{jiang2025go}, and M3ED~\cite{chaney2023m3ed} augment raw perception data with semantic segmentation labels to support terrain understanding and identifying obstacles.  

Beyond perception, other datasets focus on vehicle dynamics.  TartanDrive~\cite{triest2022tartandrive} and its successor TartanDrive 2.0~\cite{sivaprakasam2024tartandrive} provide extensive logs of wheel torque, throttle, and brake commands alongside multimodal observations for self-supervised dynamics modeling. Scaled vehicle platforms such as HOUND~\cite{talia2023demonstrating} and AutoRally~\cite{goldfain2019autorally} collect high speed off-road driving data in real-world environments to support studies of vehicle dynamics and control performance. 
However, all of these datasets are collected in uncontrolled outdoor settings, which limits the ability to deliberately adjust  environmental variables. This also increases the difficulty of obtaining accurate sensor measurements and ground truth data. For example, GPS-based localization often suffers from reduced accuracy due to signal occlusion by tree canopies and terrain structures. Another limitation is that they are not diverse enough to cover many terrain types that induce aggressive 6-DoF (Degree of Freedom) vehicle motions.

\par
To address the limitations of existing datasets, we also collect a diverse and precise dataset in Verti-Arena. Our dataset includes motion-capture-based localization along with onboard camera, IMU, and additional sensor streams to support perception, planning, control, and learning tasks in off-road autonomy research. Furthermore, the dataset contains a diverse range of driving behaviors, including aggressive maneuvers, which provide a broader data distribution for learning robust kinodynamic models.

\section{VERTI-ARENA}
Verti-Arena is an indoor testbed measuring 8
$\times$8 m that includes a wide range of semantic terrain configurations across varied elevation profiles. Each terrain type can be reconfigured on demand to produce different layouts that replicate natural off-road environments. The arena is enclosed by an eight-camera motion capture system that provides high-precision ground-truth pose and motion data. Within this testbed, we collect a comprehensive dataset by combining motion capture–based localization with synchronized onboard camera images, inertial measurements, and control signals for data-driven approaches in off-road autonomy.

\subsection{Geometry}
\begin{figure}[t]
  \centering
  \includegraphics[width=1.0\linewidth]{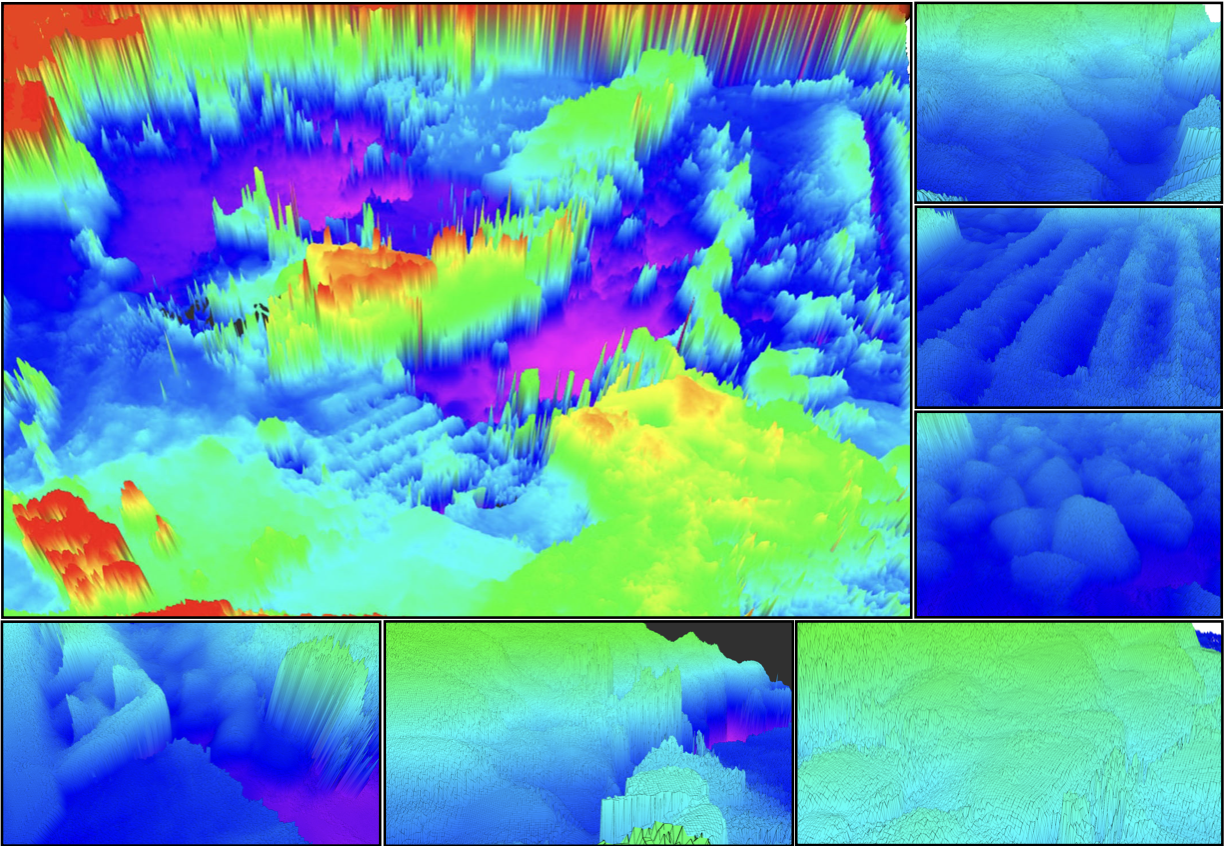}
  \caption{\textbf{Elevation Map.} The top-left region shows the complete elevation map, while the surrounding sub-images display selected detailed areas. }
  \label{fig:elevation_map}
\end{figure}
Verti-Arena features varied geometric structures that pose significant challenges for vehicle control and navigation. Specifically, gradually changing slopes cause continuously shifting load distributions. In contrast, abrupt features such as cliffs that induce rollovers or narrow crevices that suspend wheels and trap the chassis significantly increase traversal difficulty and influence route selection. To simulate these challenges, Verti-Arena incorporates elevation changes such as hills, cliffs, and ravines. Furthermore, it ensures that a direct connection between any two points within Verti-Arena is not always physically traversable by a vehicle. As a result, this environment enables evaluation of planners’ ability to generate feasible paths under these spatial constraints.
\par
To provide a clearer representation of the geometry, we use the Elevation Mapping CuPy software package~\cite{miki2022elevation, Fankhauser2014RobotCentricElevationMapping}
to generate an elevation map. As shown in Fig.~\ref{fig:elevation_map}, the terrain exhibits diverse elevation variations, with differences reaching up to 0.7 meters. It includes large-scale structures such as hills and cliffs, as well as fine-grained features like narrow trenches and bridge-like gaps, all of which introduce both global and local challenges for off-road navigation.

\par

\subsection{Semantics}
In real-world off-road scenarios, vehicles must handle not only complex spatial geometries but also rich semantic information. Verti-Arena includes diverse terrain types that reproduce the physical and perceptual characteristics of natural,  irregular, off-road environments. Each terrain category varies in deformability, surface texture, and ease of traversal, introducing challenges such as sinking, slipping, or rollover. 
\par
\begin{figure}[t]
  \centering
  \includegraphics[width=1.0\linewidth]{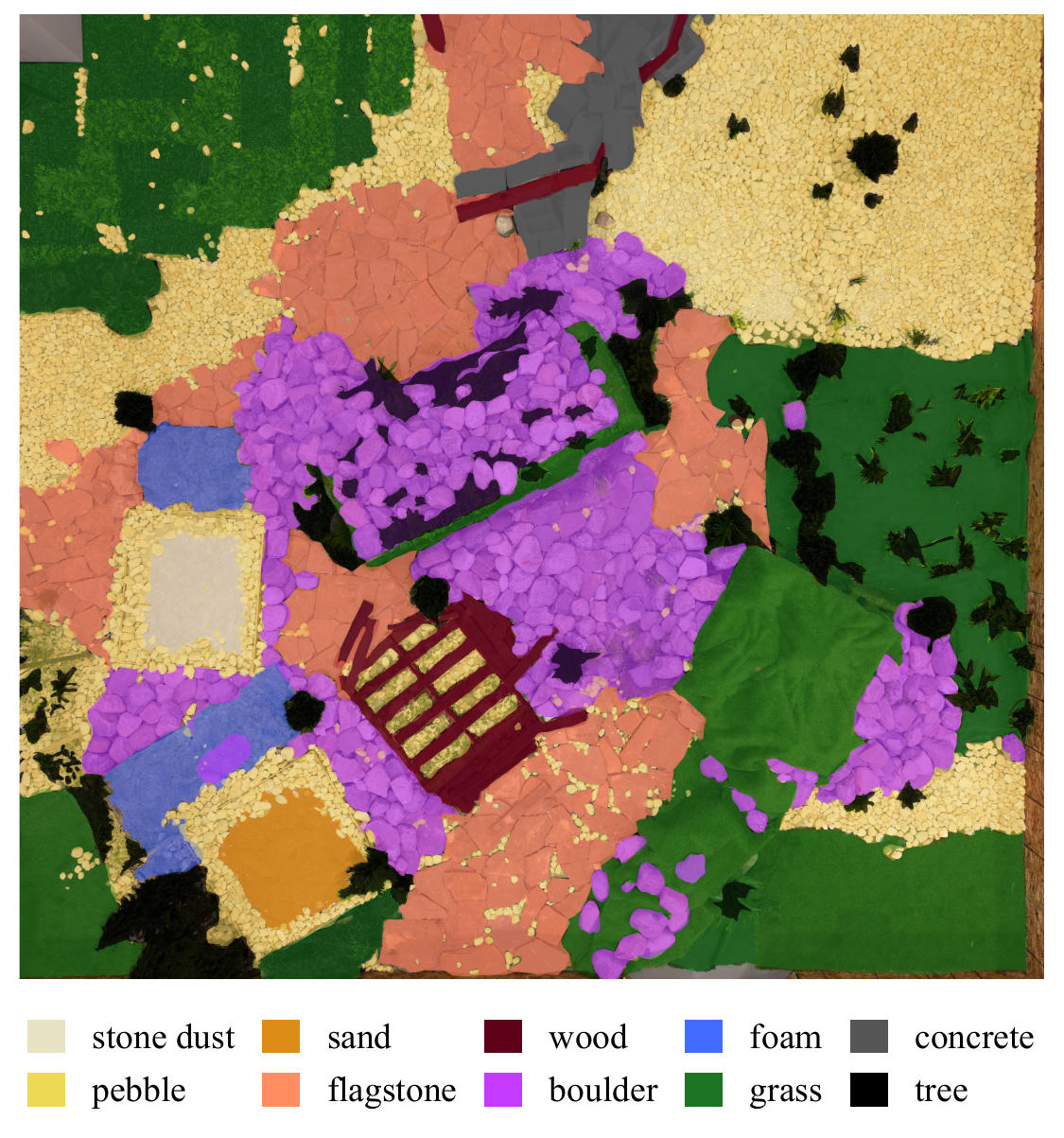}
  \caption{Semantic Map with Ten Semantic Classes Blended Together.}
  \label{fig:semantic_map}
\end{figure}
\begin{figure}[t]
  \centering
  \includegraphics[width=1.0\linewidth]{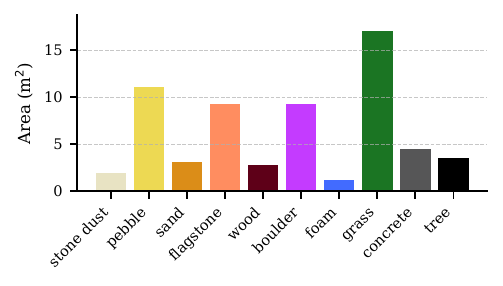}
  \caption{Semantic Distribution of Verti-Arena.}
  \label{fig:semantic_label_distribution}
\end{figure}

The testbed features three deformable surfaces: sand, stone dust, and foam board. These occupy 10.30\% of the total area.  It also includes seven rigid elements: large boulders, small pebbles, grass, flagstones, wood, concrete, and trees. The semantics of each region in the testbed are illustrated in Fig.~\ref{fig:semantic_map}, and the distribution of semantic categories is presented in Fig.~\ref{fig:semantic_label_distribution}.  Within a single category, there are significant internal variations. For example, grass patches range from closely cut turf to dense, overgrown grass, weed, and hay, while trees include both low shrubs and tall trunks.
\par
Rather than isolating each semantic type into neatly structured sections, Verti-Arena blends them to reflect the continuous variation observed in natural settings. Clusters of grass emerge between stones, wooden planks are positioned across rocky surfaces, and fine pebbles fill the gaps between larger flagstones. This blended semantic distribution ensures that no two regions are identical and creates a realistic, visually and dynamically complex environment for off-road autonomy research. 
\par

\subsection{Obstacles}
Obstacles refer to spaces or objects that a vehicle cannot navigate through due to, e.g., rough terrain, insufficient power, or limited room to maneuver. In Verti-Arena, natural obstacles include large boulders, steep hills, and certain types of vegetation. Specifically, boulders and hills can be too tall or too steep for a vehicle to pass, resulting in wheel slip, loss of traction, or even tip-over. Vegetation, such as dense shrubs, can entangle the wheels or damage onboard sensors. In addition to natural features, man-made barriers are also present. These include reinforced concrete walls that completely block the path and must be detected and avoided. Ultimately, whether something constitutes an obstacle depends on the capabilities and limitations of the vehicle and is not always clear, i.e., certain vehicles equipped with certain mobility systems may be able to negotiate through, while others may not.  These vehicle-terrain interactions can be predicted based on both semantic and geometric properties.

\subsection{Variability}

The testbed is designed to support a wide range of terrain configurations. Fig.~\ref{fig:elevation_map}  and Fig.~\ref{fig:semantic_map} only illustrate one representative example of the arrangement of terrain geometry and semantics. However, this configuration is not fixed. During experiments, terrain elements can be shuffled: stones can be repositioned, sections of grass turf may be removed, and wooden bars can be placed between mountains as bridges to form alternative traversal paths. This flexibility enables the recreation of a wide range of environments for data collection and evaluation using a diverse set of physical materials.

\begin{figure*}[t]
  \centering
\includegraphics[width=1.0\linewidth]{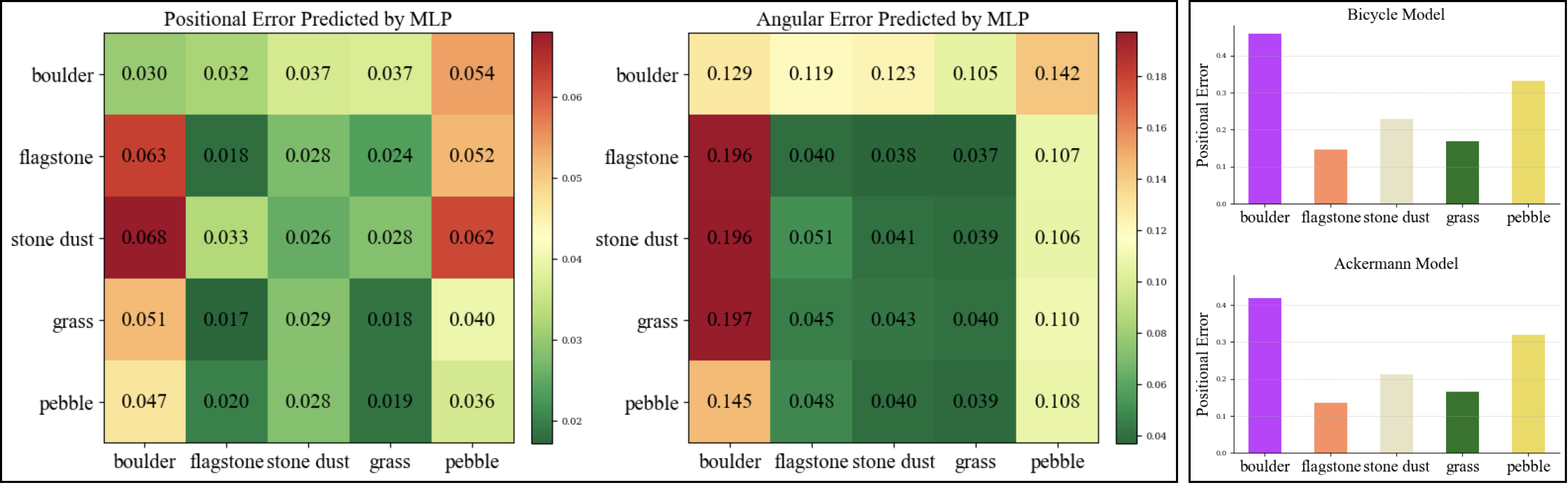}
  \caption{Positional and Angular Errors from MLPs (Left) and Positional Errors from Kinematic Models (Right).}
  \label{fig:exp_result}
\end{figure*}
\begin{figure}[t]
  \centering
  \includegraphics[width=1.0\linewidth]{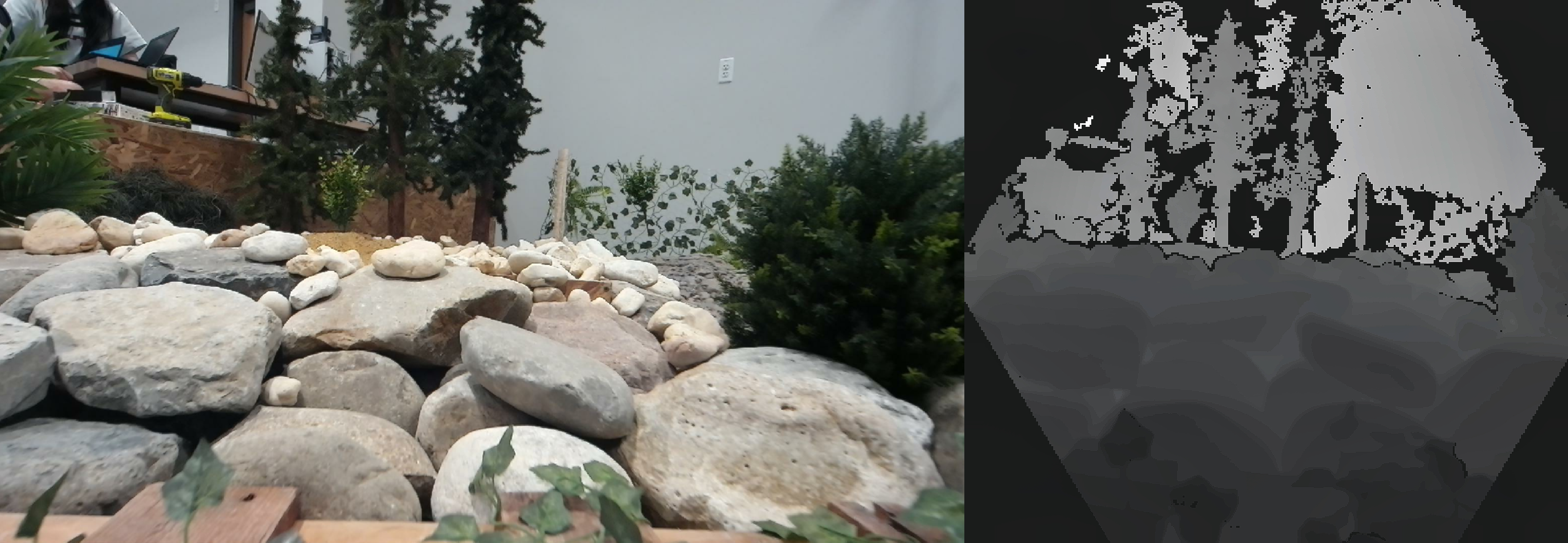}
  \caption{Example RGB (Left) and Depth (Right) Images in the Dataset.}
  \label{fig:example_img}
\end{figure}

\subsection{Datasets}
\label{sec:Datasets}
We use a four-wheeled ground vehicle (V4W; 0.523 × 0.249 × 0.20 m)~\cite{datar2024toward}, equipped with a Microsoft Azure Kinect RGB-D camera mounted on a single DoF gimbal actuated by a servo that keeps the field of view fixed on the terrain ahead, independent of chassis orientation. The platform also includes an IMU, wheel encoders, and an NVIDIA Jetson Xavier NX for onboard processing, and is time-synchronized with an eight-camera motion-capture system for data collection. We teleoperate the vehicle using a controller to navigate Verti-Arena to induce a variety of vehicle-terrain interactions with aggressive 6-DoF vehicle poses, including rollover and immobilization. 

\par
All sensors and data streams are fully integrated into the ROS 2 ecosystem. Each trajectory is stored individually as a ROS 2 bag file, resulting in a dataset comprising thousands of runs. Each raw bag file includes the following components:

\begin{itemize}
 
 \item Color and depth images:
   RGB images ($1280 \times 720 \times 3$) and depth images ($512 \times 512$) captured by onboard cameras. Intrinsic calibration parameters for both modalities are included. An example frame is shown in Fig.~\ref{fig:example_img}.

  \item Inertial measurements:  
    Gyroscope and accelerometer readings from the onboard IMU, sampled at 100 Hz.
  \item High-level control inputs:  
    Teleoperation and autonomous control commands, including differential lock status (2D binary vector for front and rear differentials), gear mode (1D binary vector indicating low or high gear), drive velocity, and steering angle.
    \item Low-level actuator feedback: Joint states obtained from the motor control unit, including motor velocities, steering angles, and positions and velocities of all wheels.
  \item Coordinate transforms:  
    Time-varying and static TF2 transform trees for all frames, including 6-DoF pose estimates of tracked rigid bodies recorded via a motion capture system, which serve as high-precision ground truth for localization and motion analysis. 
\end{itemize}

\section{EVALUATION AND DISCUSSIONS}
To quantitatively validate the terrain diversity in Verti-Arena, we assess how vehicle kinodynamics varies across distinct zones of the testbed. We select five representative terrain types from Verti-Arena: boulder, flagstone, stone dust, grass, and pebble.
For each zone, a separate dataset of vehicle trajectories is collected, providing the complete state transitions of the vehicle over time.
\par

To analyze and compare the underlying dynamics across different zones, we formalize the system’s behavior as a forward model:
\begin{equation}
    x_{t+1} = f(x_t, u_t),
    \nonumber
\end{equation}
where $x_t \in \mathcal{X}$ denotes the vehicle’s state at time $t$, $u_t \in \mathcal{U} \subset \mathbb{R}^2$  represents the throttle and steering control input.
All data are collected in a fixed low-gear driving mode with locked differentials. Furthermore, all vehicle states are transformed into the vehicle’s body frame. Notice that we intentionally omit the environmental features in the forward model input to highlight the differences caused by them. 
\par
 We evaluate three classes of predictive models: a bicycle model, an Ackermann model, and a multilayer perceptron (MLP). The bicycle and Ackermann models are tuned using the vehicle’s actual physical parameters.  For the MLP, a separate model instance is trained and tested independently for each terrain zone using the corresponding data.

The heatmaps in Fig.~\ref{fig:exp_result} (left, middle) exhibit a clear diagonal pattern: MLP models perform best when evaluated on the same terrain zone used for training, suggesting a specialization to zone-specific dynamics. The variation in prediction errors across terrain zones further indicates that vehicle kinodynamics are terrain-dependent. Moreover, cross-zone transfer is better facilitated between more similar terrains. For example, models trained on flagstone transfer to grass (and vice versa) with relatively small errors, implying that generalization deteriorates as terrain dissimilarity increases.

In contrast, the bar plots in Fig.~\ref{fig:exp_result} (right) highlight the limitations of classical kinematic baselines. The bicycle and Ackermann models, even when carefully calibrated to the vehicle’s physical parameters, consistently yield larger positional errors. Their planar, low-slip assumptions fail to capture the full 6-DoF motion, including pitch and roll dynamics, transient wheel lift, and tire–terrain interactions involving slip and compliance. This limitation results in particularly poor performance on the pebble and boulder terrains.

\par
Beyond model specialization, the results also reveal clear differences in terrain difficulty. Based on model performance, the terrain zones can be ranked in decreasing order of difficulty: boulder, pebble, stone dust, grass, and flagstone. Among them, boulder and pebble are the most challenging. The boulder surface exhibits large height variations and intermittent wheel–ground contact, which cause significant pitch and roll motion as well as frequent wheel lift. The pebble terrain consists of small loose stones that roll under load, leading to continuous slip and occasional sinkage. As a result, models tested on pebbles or boulders exhibit higher positional and angular errors, regardless of training terrain or model type. In contrast, grass and flagstone provide relatively smooth, continuous support with high friction, resulting in consistently lower errors across models and aligning with the above ranking.

We note several limitations of this evaluation. In particular, the stone dust zone is physically small, which constrains the driving paths and yields a relatively narrow distribution of states and commands. This sampling bias likely makes the data from stone dust more homogeneous than those of larger zones. Lower errors on this surface therefore do not imply that stone dust is inherently easy in natural environments, where deformable effects such as sinkage are more obvious. Moreover, all results are obtained in a fixed low-gear, locked-differential configuration. Performance at higher speeds with unlocked differentials may differ.

\section{CONCLUSIONS}
We present Verti–Arena, a controllable and reconfigurable indoor testbed that integrates ten semantic terrain types with diverse elevations, instrumented with motion-capture ground truth and synchronized onboard sensing. We provide standardized datasets and a web-based interface for remote experimentation, enabling the quantification of off-road mobility and data-driven autonomy. Experiments across five representative terrains demonstrate that vehicle kinodynamics are terrain-dependent and that terrain difficulty varies substantially across surfaces, highlighting the need for terrain-aware modeling in off-road autonomy.

\section{ACKNOWLEDGEMENTS}
This work has taken place in the RobotiXX Laboratory at George Mason University. RobotiXX research is supported by National Science Foundation (NSF, 2350352), Army Research Office (ARO, W911NF2320004, W911NF2420027, W911NF2520011), Air Force Research Laboratory (AFRL), US Air Forces Central (AFCENT), Google DeepMind (GDM), Clearpath Robotics, Raytheon Technologies (RTX), Tangenta, Mason Innovation Exchange (MIX), and Walmart. The work of F. Cancelliere has been supported by the European Union - Next Generation EU, Mission 4 Component 2 Line 1.3, under the PNRR MUR project PE0000013 - FAIR "Future Artificial Intelligence Research".

\addtolength{\textheight}{-10.5cm}   % This command serves to balance the column lengths

%%%%%%%%%%%%%%%%%%%%%%%%%%%%%%%%%%%%%%%%%%%%%%%%%%%%%%%%%%%%%%%%%%%%%%%%%%%%%%%%

%%%%%%%%%%%%%%%%%%%%%%%%%%%%%%%%%%%%%%%%%%%%%%%%%%%%%%%%%%%%%%%%%%%%%%%%%%%%%%%%



% \section*{ACKNOWLEDGMENT}

% The preferred spelling of the word ÒacknowledgmentÓ in America is without an ÒeÓ after the ÒgÓ. Avoid the stilted expression, ÒOne of us (R. B. G.) thanks . . .Ó  Instead, try ÒR. B. G. thanksÓ. Put sponsor acknowledgments in the unnumbered footnote on the first page.

%%%%%%%%%%%%%%%%%%%%%%%%%%%%%%%%%%%%%%%%%%%%%%%%%%%%%%%%%%%%%%%%%%%%%%%%%%%%%%%%

\bibliographystyle{IEEEtran}
\bibliography{references}

\end{document}